\documentclass{article}
\usepackage{ijcai20}

\usepackage{times}
\usepackage{soul}
\usepackage{url}
\usepackage[hidelinks]{hyperref}
\usepackage[utf8]{inputenc}
\usepackage[small]{caption}
\usepackage{graphicx}
\usepackage{amsmath}
\usepackage{amsthm}
\usepackage{booktabs}
\usepackage{xcolor}
\usepackage{makecell}
\usepackage[ruled, vlined, linesnumbered]{algorithm2e}
\title{Investigating the Decoders of Maximum Likelihood Sequence Models: \newline A Look-ahead Approach}

\author{
Yu-Siang Wang$^1$
\and
Yen-Ling Kuo$^2$\and
Boris Katz$^2$
\affiliations
$^1$University of Toronto, $^2$MIT
\emails
yswang@cs.toronto.edu,
\{ylkuo, boris\}@mit.edu
}

\begin{document}

\maketitle

\begin{abstract}




We demonstrate how we can practically incorporate multi-step future information into a decoder of maximum likelihood sequence models.
We propose a ``$k$-step look-ahead'' module to consider the likelihood information of a rollout up to $k$ steps.
Unlike other approaches that need to train another value network to evaluate the rollouts, we can directly apply this look-ahead module to improve the decoding of any sequence model trained in a maximum likelihood framework.
We evaluate our look-ahead module on three datasets of varying difficulties: IM2LATEX-100k OCR image to LaTeX, WMT16 multimodal machine translation, and WMT14 machine translation.
Our look-ahead module improves the performance of the simpler datasets such as IM2LATEX-100k and WMT16 multimodal machine translation.
However, the improvement of the more difficult dataset (e.g., containing longer sequences), WMT14 machine translation, becomes marginal.
Our further investigation using the $k$-step look-ahead suggests that the more difficult tasks suffer from the overestimated EOS (end-of-sentence) probability.
We argue that the overestimated EOS probability also causes the decreased performance of beam search when increasing its beam width.
We tackle the EOS problem by integrating an auxiliary EOS loss into the training to estimate if the model should emit EOS or other words.
Our experiments show that improving EOS estimation not only increases the performance of our proposed look-ahead module but also the robustness of the beam search. 

\end{abstract}

\section{Introduction}
\begin{figure}
    \centering
    \includegraphics[width = 0.45\textwidth]{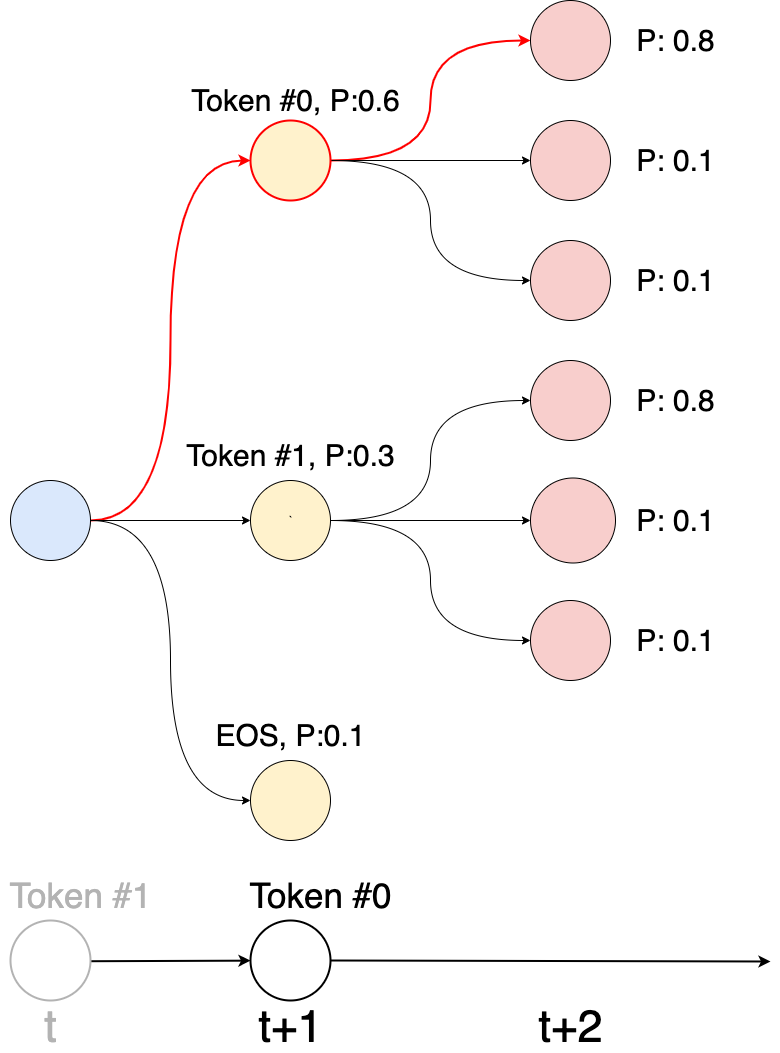}
    \caption{A synthetic example to illustrate a $2$-step look-ahead inference module for the decoder. Each node represents a word token and a probability value from the decoder. The model is predicting the word at time step $t+1$. The vocabulary size $|V|$ in this example is set to three and consists of three tokens \{Token \#0, Token \#1, and EOS(end-of-sentence)\}. We don't expand the tree from the EOS node because the node implies the end of the sentence. The depth of the expanded tree is 2 in the $2$-step look-ahead scenario. When we predict the word at time step $t$, we compute the summation of the log probabilities from the node at time step $t+1$ to the leaf of the tree. We select the word which has the maximum summation of the log probabilities along its path as our prediction at time step $t+1$. In the example, the Token \#0 has the maximum likelihood (0.8$\cdot$0.6) among the entire paths from $t+1$ to $t+2$. So we choose Token \#0 as the prediction at time step $t+1$.
    }
    \label{fig:dfs}
\end{figure}

Neural sequence models~\cite{rnn,lstm,transformer} have been widely applied to solve various sequence generation tasks including machine translation~\cite{DBLP:journals/corr/BahdanauCB14}, optical character recognition~\cite{latexocr}, image captioning~\cite{DBLP:conf/cvpr/VinyalsTBE15,DBLP:conf/cvpr/ZhouZCSX18}, visual question answering~\cite{DBLP:conf/iccv/AntolALMBZP15,DBLP:conf/emnlp/LeiYBB18}, and dialogue generation~\cite{DBLP:conf/emnlp/LiMSJRJ17}.
Such neural-based architectures model the conditional probability $P(\mathbf{y}|\mathbf{x})$ of an output sequence $\mathbf{y}=\{y_{1}, y_{2},...,y_{T}\}$ given an input $\mathbf{x}=\{x_{1}, x_{2},...,x_{N}\}$.
By using these neural models, sequence decoding can be performed by Maximum a \textit{posteriori} (MAP) estimation of the word sequence, given a trained neural sequence model and an observed input sequence.
However, in settings where the vocabulary size is huge and the length of the predicted sequence is long, the exact MAP inference is not feasible.
For example, a size $V(>=10000)$ vocabulary and a length $T(>=30)$ target sequence would lead to $V^{T}$ total possible sequences.
Other approximate inference strategies are more commonly used to decode the sequences than exact MAP inference.

The most simple decoding strategy is to always choose the word with the highest probability at each time step.
This greedy approach doesn't give us the most likely sequence and is prone to have grammatical errors in the output sequence.
Beam search (BS), on the other hand, maintains $\beta$ top-scoring successors at each time step and then scores all expanded sequences to choose one to output.
BS decoding strategy has shown good results in many sequence generation tasks and has been the most popular decoding strategy so far.
Although the beam search considers the whole sequence for scoring, it only uses the current node to decide the nodes to expand and doesn't consider the possible future to expand a node.
This incompleteness in the search leads to sub-optimal results.

When speaking or writing, we do not just consider the last word we generate to choose the next word; we also consider what we want to say or write in the future.
Regarding the future output is crucial to improve sequence generation.
For example, Monte-Carlo Tree Search (MCTS) focuses on the analysis of promising moves and has achieved great success in game-play, e.g., Go~\cite{alphago}.
An MCTS-based strategy predicts the next action by carrying out several rollouts from the present time step and calculate the reward for each rollout using a trained value network.
It then makes the decision at each time step by choosing the action which leads to the highest average future rewards.
However, MCTS requires to train another value network and takes more runtime to run the simulation.
It is not practical to run MCTS to decode sequences with large vocabulary size and many time steps.
Instead of applying MCTS in sequence decoding, we propose a new $k$-step look-ahead ($k$-LA) module that doesn't need an external value network and has a practical run-time.
Namely, our proposed look-ahead module can be plugged into the decoding phase of any existing sequence model to improve the inference results.

Figure~\ref{fig:dfs} illustrates how $k$-LA works in a $k=2$ example to choose a word at time step $t+1$.
At time step $t$, we expand every word until the search tree reaches time step $t+k$.
For each word in the tree rooted at time $t$, we can compute the likelihood of extending that word from its parent using the pretrained sequence model.
To select a word at $t+1$, we choose the word whose sub-tree has the highest accumulated probability, i.e., the highest expected likelihood in the future.

We test the proposed $k$-step look-ahead module on three datasets of increasing difficulties: IM2LATEX OCR, WMT16 multimodal English-German translation, and WMT14 English-German machine translation.
Our results show that the look-ahead module can improve the decoding in IM2LATEX and WMT16 but only marginal over the greedy search in WMT14.
Our analysis suggests that the more difficult datasets (usually containing longer sequences, e.g., WMT14 and a subset of WMT16 where sequence length $\geq25$) suffer from the overestimated end-of-sentence (EOS) probability.
The overestimated EOS probability encourages the sequence decoder to favor short sequences.
Even with the look-ahead module, the decoder still cannot recover from that bias.
To fix the EOS problem, we use an auxiliary EOS loss in training to make a more accurate EOS estimation.
We show that the model trained with the auxiliary EOS loss not only improves the performance of the look-ahead module but also makes the beam search more robust.


This work makes a number of contributions.
We show how we can incorporate future information to improve the decoders using pretrained sequence models only.
Our analysis with the proposed decoder also help us pinpoint the issues of the pretrained sequence model and further fix the sequence model.
We expect that looking into both decoders and models together can provide a better picture of sequence generation results and help design a more robust sequence model and training framework.

\begin{figure*}
    \centering
    \includegraphics[width=0.75\textwidth]{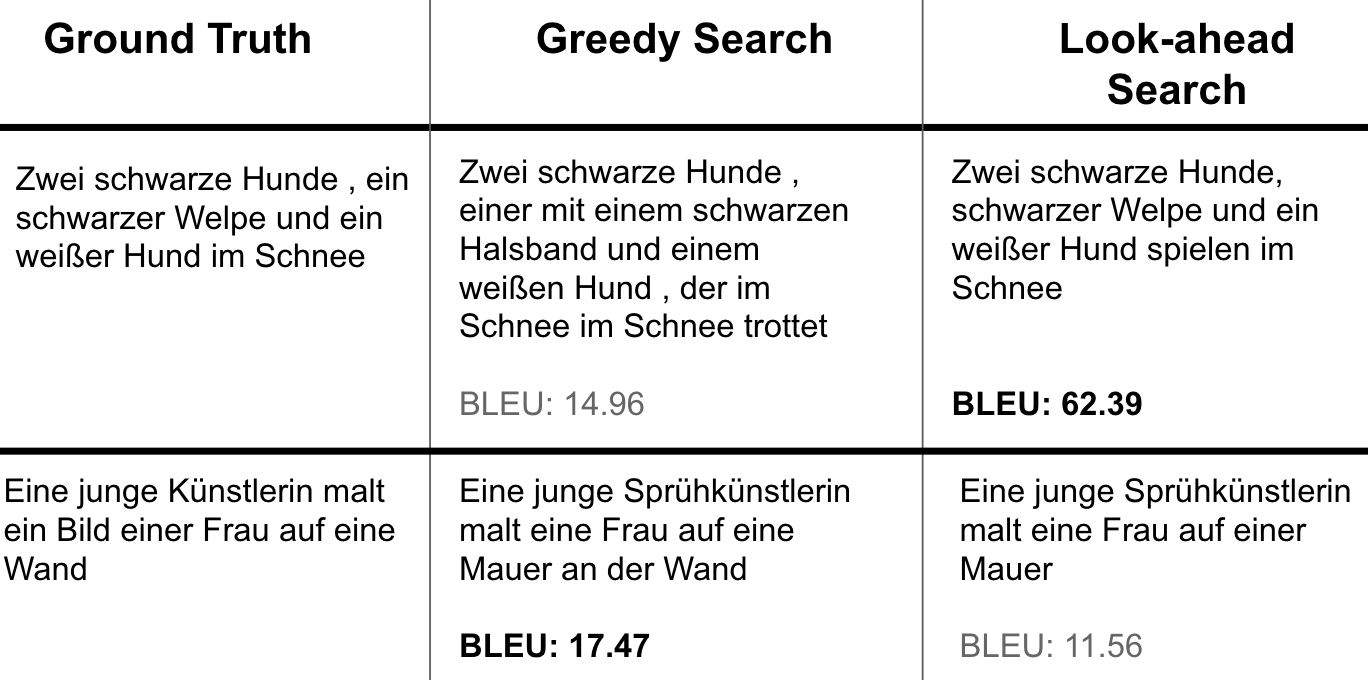}
    \caption{Two examples from the WMT16 dataset. The input of the first row English sentence is, ``Two black dogs , a black puppy , and a white dog in the snow" and the input of the second row English sentence is ``A young female artists paints an image of a woman on a wall". We exhibit the translation results with different strategies. The first row illustrates the successful example and the second row shows the fail example of the look-ahead module. We define the successful example in terms of the BLEU scores.}
    \label{fig:viz_la}
\end{figure*}

\section{Related Work}

\textbf{Learning to search with look-ahead cues}:
Reinforcement learning (RL) techniques, especially the value network, are often used to incorporate hypothetical future information into predictions.
\cite{DBLP:journals/corr/BahdanauCB14} train their policy and value networks by RL but allow the value network to also take the correct output as its input so that the policy can optimize for BLEU scores directly.
\cite{DBLP:conf/cvpr/ZhouZCSX18} train image captioning policy and value networks using actor-critic methods.
The authors found that the global guidance introduced by the value network greatly improves performance over the approach that only uses a policy network.
\cite{alphago} apply self-play and MCTS to train policy and value networks for Go.
It show that MCTS is a powerful policy evaluation method.
\newline
\textbf{Augmenting information in training sequence model}:
\cite{rewardmle} focus on using auxiliary reward to improve the maximum likelihood training decoder.
They define the auxiliary reward as the negative edit distance between the predicted sentences and the ground truth labels.
\cite{OCD} optimize the seq2seq models based on edit distance instead of maximizing the likelihood.
They show the improvements on the speech recognition dataset. 
\cite{DBLP:conf/emnlp/WisemanR16} focus on improving the decoder by alleviating the mismatch between training and testing.
They introduced a search-based loss that directly optimizes the network for beam search decoding.
\newline
\textbf{Sequence modeling errors}:
\cite{nmterror} analyze the machine translation decoder by enumerating all the possible predicted sequences. 
They predict the decoded sequence by choosing the sequence with the highest likelihood.
Their results demonstrate that the neural machine translation model usually assigned its best score to the empty sentences for over 50\% of inference sentences.
In \cite{DBLP:conf/interspeech/ChorowskiJ17}, they argue that the seq2seq models suffer from the overestimated word probability in the training stage. 
They propose to solve the issue using the label smoothing technique.

\section{Datasets}

In the following discussions of the paper, we evaluated the proposed approaches on three different datasets: IM2LATEX-100k OCR dataset~\cite{latexocr}, WMT16 Multimodal English-German (EN-DE) machine translation dataset~\cite{multi30k}, and WMT14 English-German (EN-DE) machine translation dataset.
In IM2LATEX-100K, the input is given an image and the goal is to generate the corresponded LaTeX equation.
The dataset is separated into the training set (83,883 equations), the validation set (9,319 equations) and the test set (10,354 equations).
The average length of the target LaTeX equations is 64.86 characters per equation.
The WMT16 multimodal dataset consists of 29,000 EN-DE training pairs, 1,014 validation pairs and 1,000 testing pairs. 
Each EN-DE training pair are the descriptions of an image.
The average length of the testing target sentences is 12.39 words per sentence.
In this paper, we didn't use the support from the image information.
The WMT14 EN-DE machine translation dataset consists of 4,542,486 training pairs, 1,014 validation pars.
We train on WMT14 data but evaluate the model on the newstest2017 dataset which consists of 3,004 testing pairs.
The average length of target sequences in newstest2017 is 28.23 words per sentence, which is a much longer compared to the average length of the WMT16 translation dataset.
The longer target sequences in WMT14 makes WMT14 a more difficult task than the WMT16 translation task.

\begin{algorithm}[t]
\caption{DFS Look Ahead for Prediction}
\label{alg: dfs}
\DontPrintSemicolon
\textbf{Input}: Pretrained sequence model $f_\theta$ parameterized by $\theta$, Max time step $T$, look ahead step $k$,  \\
Initialize predicted sentence $S$ to empty sequence \\
Initialize $w_{0}$ as the $<BOS>$ \\
$<BOS>$: Begin-of-Sentence token\\
Initialize max\_prob to -INF

\SetKwFunction{FMain}{DFSLookAhead}
  \SetKwProg{Fn}{Function}{:}{}
  \Fn{\FMain{$probs$, $words$, $t\_dfs$, $cum\_prob$, $head$}}{
        \For{prob, word in (probs, words)}{
            $cum\_prob$ += prob \\
            \If{$cum\_prob$ $<$ max\_prob}{
                break
            }
            \If{$t\_dfs == k$ or word == $<EOS>$}{
                break \\
                \If{$cum\_prob$ $>$ max\_prob}{
                    max\_prob = $cum\_prob$ \\
                    $w_{t}$ = head
                }
            }
            \Else{
                $probs$ = $f_{\theta}(w_{t\_dfs-1}, h_{t\_dfs})$ \\
                $probs$, $words$ = Sorted($probs$) \\
                \If{$t\_dfs$ == 1}{
                    $head$ = word
                }
                DFS($probs$, $words$, $t\_dfs$+1, $cum\_prob$, $head$)
            }
        }
        \KwRet\;
  }
\For{$t \gets 1$ to $T$ }{
    Initialize $t\_dfs$ to 1 \\
    Initialize $cum\_prob$ to 0 \\
    Initialize $w_{t}$ to None \\
    Initialize $head$ to None \\
    $probs$ = $Decoder(w_{t-1}, h_{t})$ \\
    $probs$, $words$ = Sorted($probs$) \\
    DFSLookAhead($probs$, $words$, $t\_dfs$, $cum\_prob$, $head$) \\
    S.append($w_{t}$) \\
}
\end{algorithm}

\section{Look-ahead Prediction}
We present a look-ahead prediction module to take advantage of the future cues.
This proposed look-ahead moduel is based on depth-first search (DFS) instead of using the Monte-Carlo Tree-based (MCTS) method.
In the DFS-based look-ahead module, we are able to prune the negligible paths and nodes whose probability is too small to be the word lead to the largest probability.
In contrast, MCTS-based method requires plenty of samples to estimate the nodes' expected probability.
To compare the real execution of these two look-ahead methods, we test both methods on the transformer model trained on the WMT14 dataset.
We run the experiment on Tesla V100 GPU with 500 input sentences.
We set the look-ahead time step equals to 3 for both search strategies.
In the MCTS setting, we operate 20 rollouts in each time step and the average execution time is 32.47 seconds per sentence.
As for DFS-based method, the average execution time is 0.60 seconds per sentence.
To make the look-ahead module more practical, we choose the DFS-based look-ahead module as our node expansion strategy.

\subsection{Method}
Figure \ref{fig:dfs} illustrate our proposed DFS look-ahead module.
Algorithm \ref{alg: dfs} is the pseudo-code of the proposed method.
Given a pretrained sequence model and a size $|V|$ vocabulary, we are able to expand a tree in the current time step $t$ to the $t+k$ in the $k$-step look-ahead setting.
The height of the tree is k.
For example, in the $2$-step look-ahead setting, there are $O(|V|)$ nodes at height 1 and $O(|V|^{2})$ leaf nodes at height 2.
At $t+1$, we select the word which has the maximum summation of the log-likelihood along the path from height 1 to the leaf nodes.
We repeat the previous operation at each time step until we predict the EOS token.
Although the time complexity of the DFS is $O(|V|^{k+1})$, we are able to prune a lot of insignificant paths in our tree.
At line 9 in Algorithm \ref{alg: dfs}, we early stop DFS when then current cumulative log-probability is smaller than the maximum summation of log-probability we have encountered so far.
Since we sort log probabilities before we perform the DFS, we are able to prune many paths which can't be the optimal path in the expanded tree.
By using the foresight word information in the prediction, we can select the word guiding to the largest probability in advance. 

\subsection{Experiments}
We train and test the sequence models using OpenNMT \cite{2017opennmt}.
For the IM2LATEX-100K image to LaTeX OCR dataset, our CNN feature extractor is based on \cite{DBLP:conf/acl/GehringAGD17} and we pass the visual features in each time step to a 512 hidden units bi-LSTM model.
For the WMT16 EN-DE translation dataset, we trained an LSTM model with 500 hidden units.
As for the WMT14 EN-DE translation dataset, we trained a transformer model with 8 heads, 6 layers, 512 hidden units and 2048 units in the feed-forward network.
We report the BLEU scores of the greedy search and the look-ahead (LA) search with different $k$-steps in all three datasets.
In our look-ahead module definition, the 1-LA setting is equivalent to the greedy search since we only use the current time step information.
The look-ahead module is more directly comparable to the greedy search method than the beam search method because the beam size of either the greedy search or the look-ahead module is 1.
For a better reference of the range of the performance, we also report the scores of the beam search.
Note that we may combine the beam search method and the look-ahead method at the same time.
For simplicity, we test our look-ahead module with the beam width = 1 setting.

\subsection{Results}
We test the look-ahead module with five different settings, which are 1-LA (Greedy) to 5-LA and we evaluate the models with Sacre BLEU scores~\cite{post-2018-call} which is a commonly used machine translation metric.
We demonstrate the results of three different models in Table \ref{tab:ori_la_im2latex}, \ref{tab:ori_la_multi30k}, and \ref{tab:ori_la_mt17}. 
Our results show that the look-ahead module can improve the models on the IM2LATEX-100K dataset and the WMT16 dataset. We show the examples of using the look-ahead module on the model trained on the WMT16 dataset in Figure \ref{fig:viz_la}. However, the improvement becomes marginal on the WMT14 dataset and even harms the performances in the 5-LA setting. We argue that the look-ahead module might be less effective on the more difficult datasets, i.e., the longer target sequences. We show that in Table \ref{tab:ori_la_multi30k}, both the look-ahead module and the beam search harm the model on the target sequences longer than 25 words. We didn't discuss IM2LATEX task because the accuracy of IM2LATEX task is highly dependent on the recognition accuracy of the CNN models and this makes the model a different scheme compared to the rest of the two textual translation models. We argue that the ineffectiveness of the look-ahead module on WMT14 is caused by the overestimated end-of-sentence (EOS) probability. The overestimated EOS probability will lead to shorter sentences and make the wrong prediction at the same time. 

To support our argument, we show the average length differences between the predicted sequences and the ground truth sequences. For each sentence, the difference is calculate by (Prediction Length - Ground Truth Length). Therefore, a positive number indicates that the model tends to predict longer sentences than the ground truth sentences and vice versa. We test the WMT16 LSTM model and the WMT14 transformer model with different search strategies. The two trends shown in the two figures are the same. Both models tend to predict shorter sequences with the increasing of the look-ahead steps. However, the WMT16 LSTM model tend to predict ``overlong" sentences while the WMT14 transformer model usually predicts ``overshort" sentences in the greedy search setting. These two properties make the look-ahead module substantially improve the WMT16 model but marginally improve the WMT14 model. The results substantiate our argument of the overestimated EOS problem in the more difficult dataset.

In \cite{nmterror}, they enumerate all the possible sequences and find that the model assigns the highest probability to the empty sequence for over 50\% of testing sentences.
Their result is consistent with our analysis.
Both demonstrate the EOS problem in different schemes. 
However, their experiment settings are not practical because enumerating all the possible sequences in the exponential-growth search space is time-consuming. 

\begin{table}[h]
    \centering
    \begin{tabular}{c|c}
        Search Strategy & BLEU \\
        \Xhline{4\arrayrulewidth}
        Greedy Search & 86.24 \\
        \hline
        2-LA & 86.65 \\
         
        3-LA  & 86.71 \\
         
        4-LA  & 86.77 \\
         
        5-LA  & \textbf{86.79} \\
         \Xhline{2\arrayrulewidth}
         Beam Search (B=10) & 86.28 \\
        \hline
    \end{tabular}
    \caption{The performances of the IM2LATEX-100K Bi-LSTM model. We discover that the look-ahead improves the model from the greedy search method — noted that LA is more directly comparable to the greedy search because of their same beam size. We also show the scores of the beam search for the reference}
    \label{tab:ori_la_im2latex}
\end{table}

\begin{table}[h]
    \centering
    \begin{tabular}{c|c|c}
        Search Strategy & BLEU  & BLEU (Target len$\geq25$)\\
        \Xhline{4\arrayrulewidth}
        Greedy Search & 31.67 & \textbf{23.86}\\
        \hline
        
        2-LA & 32.07 & 21.50\\
         
        3-LA  & 32.20 & 22.78\\
         
        4-LA & \textbf{32.42} & 22.45\\
         
        5-LA  & 32.41 & 23.30\\
        \Xhline{2\arrayrulewidth}
        Beam Search (B=10) & 33.83 & 22.45 \\
        \hline
    \end{tabular}
    \caption{The performances of the LSTM model trained on the WMT16 multimodal translation dataset with different LA steps. We show the look-ahead module is able to improve the model on the entire testing set. However, either the LA module or the beam search method harm the models when the length of the target sentences is longer than 25 words.}
    \label{tab:ori_la_multi30k}
\end{table}

\begin{table}[h]
    \centering
    \begin{tabular}{c|c}
        Search Strategy & BLEU \\
        \Xhline{4\arrayrulewidth}
        Greedy Search & 27.50 \\
        \hline
        
        2-LA & \textbf{27.71} \\
         
        3-LA & 27.62 \\
         
        4-LA & 27.56 \\
         
        5-LA & 27.35 \\
         \Xhline{2\arrayrulewidth}
         Beam Search (B=10) & 28.21 \\
        \hline
    \end{tabular}
    \caption{We show the results of applying LA module to the transformer model trained on the WMT14 dataset. We find that the LA module slightly improves the original model but harms the performance when the LA time step is 5. We suggest one of the reasons of these results are caused by the EOS problem.}
    \label{tab:ori_la_mt17}
\end{table}

\begin{figure}[h]
    \centering
    \includegraphics[width=0.45\textwidth]{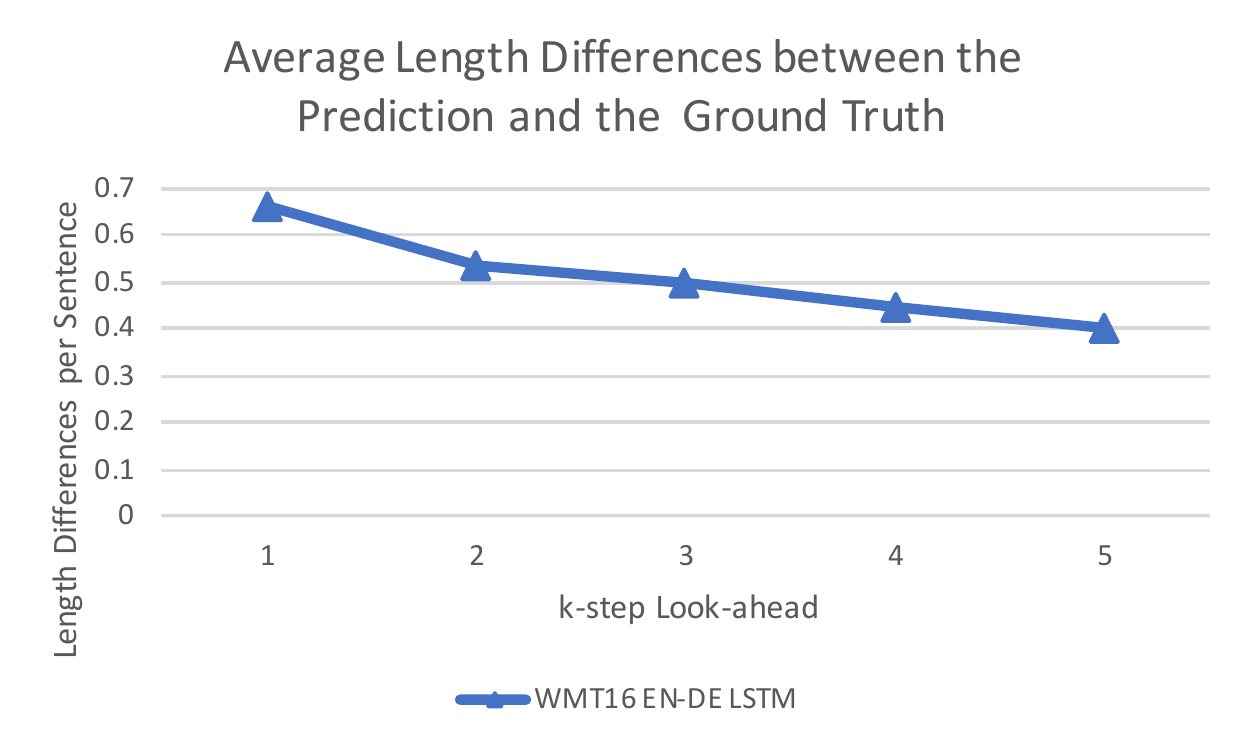}
    \includegraphics[width=0.45\textwidth]{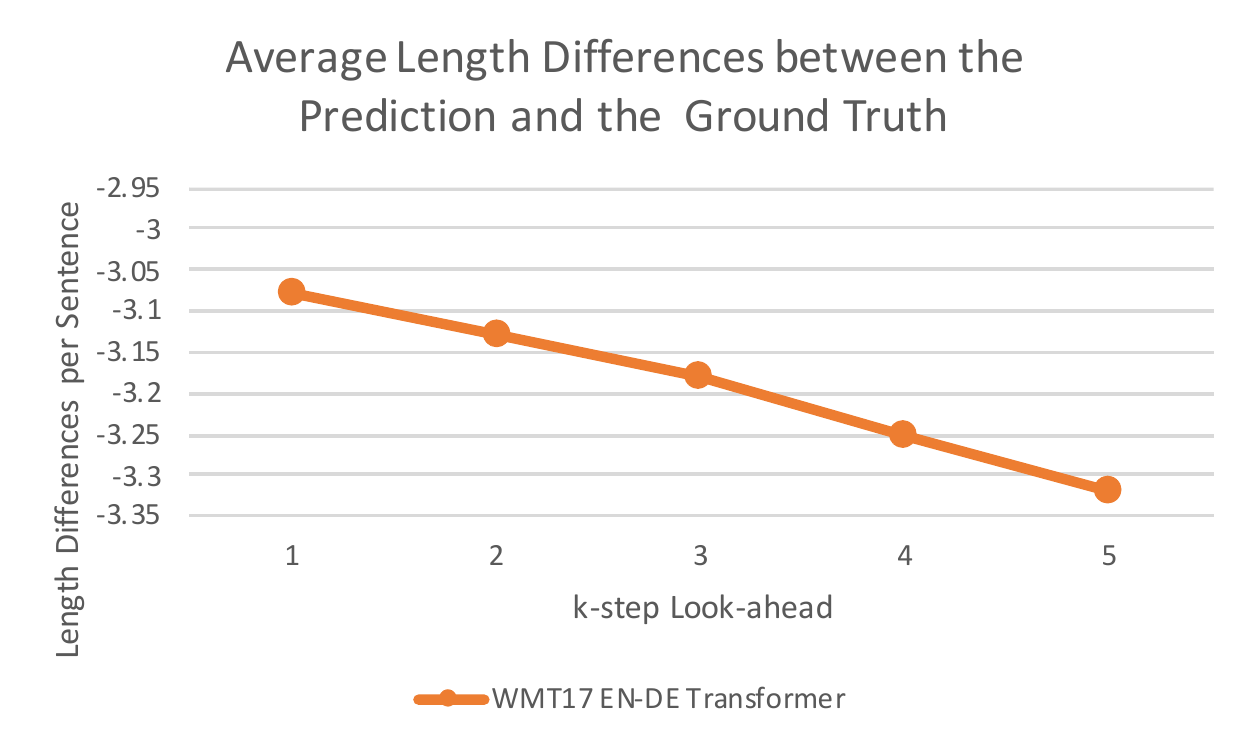}
    \caption{We demonstrated the average length differences between the predicted sequences and the ground truth sequences. A positive number means the model tends to predict longer sentences than the ground truth sentences and vice versa. With the increase of the look-ahead steps, the two models tend to predict shorter sequences.}
    \label{fig:ori_mt_eos_len}
\end{figure}

\begin{table*}[!ht]
    \centering
    \begin{tabular}{c|cccccc}
        & \multicolumn{6}{c}{$\gamma$} \\
        \hline
        Search Strategy     & 0.0     & 0.25  & 0.50  & 0.75  & 1.0   & 1.25 \\
        \Xhline{4\arrayrulewidth}
        Greedy & 27.50 & 27.81 & 27.74 & 27.75 & \textbf{27.90} & 27.71 \\
        2-LA   & 27.71 & 28.05 & 27.95 & 27.99 & \textbf{28.20} & 27.85 \\
        3-LA   & 27.89 & 27.82 & 27.87 & 27.82 & \textbf{28.10} & 27.68 \\
        4-LA   & 27.56 & 27.81 & \textbf{27.87} & 27.74 & 27.84 & 27.68 \\
        5-LA   & 27.35 & 27.71 & 27.74 & 27.63 & \textbf{27.87} & 27.55 \\

    \end{tabular}
    \caption{We show the results of integrating auxiliary EOS loss into the training state. $\gamma$ is the weight of the auxiliary EOS loss. We find the EOS loss not only boosts the performance of the model when using the greedy search, the model is more robust to the larger Look-ahead steps with reasonable weights of auxiliary EOS loss.}
    \label{tab:aux_la_mt17}
\end{table*}

\section{Auxiliary EOS Loss}\label{sec: aux_eos_loss}
To tackle the EOS problem, we introduce an auxiliary EOS loss to effectively solve the problem.
We test the model trained with our proposed auxiliary EOS loss in our proposed DFS based look-ahead setting which is more practical in the real world.

\subsection{Methods}
We ensure that the model doesn't ignore the EOS probability of the negative EOS ground truth token in each time step, i.e., the ground truth word which is not the EOS token. By given a batch of training data, the original sequence modeling loss can be represented as
\begin{equation*}
    L_{original} = \sum_{i=1}^{N}-log(P_{c})
\end{equation*}
where N is the batch size and c is the correct class of the $i^{th}$ data in the batch. We could see the original loss $L_{original}$ only focuses on the loss of the correct classes. In order to incorporate the EOS token loss into our train, we treat the auxiliary EOS task as a binary classification problem. Our auxiliary EOS loss can be written as
\begin{equation}\label{eq: eos}
    L_{EOS} = \sum_{i=1}^{N}-log(1-P_{EOS})\cdot1(y_{c} \neq y_{EOS})
\end{equation}
\begin{equation*}
    L_{final} = L_{original} + \gamma L_{EOS}
\end{equation*}
where gamma is a scalar indicating the portion of the EOS loss.

\subsection{Experiments}
 We integrate the EOS loss into the training stage of the transformer model trained on WMT14 machine translation dataset. We train the transformer model with different weights of the auxiliary EOS loss ranged from $\gamma=0.25$ to $\gamma=1.25$ and we compare the models trained with the auxiliary results with the performance of the original model ($\gamma=0.0$) under the greedy search and look-ahead search strategies. Moreover, we test the models by utilizing the beam search as the search strategy since people sometimes find that the larger beam size would seriously harm the performances. We suspect the larger beam size issue is also related to the EOS problem. To see the effectiveness of the EOS loss, we also show the average length difference of the model trained with the auxiliary EOS loss.

\subsection{Results}
\begin{figure}[h]
    \centering
    \includegraphics[width=0.4\textwidth]{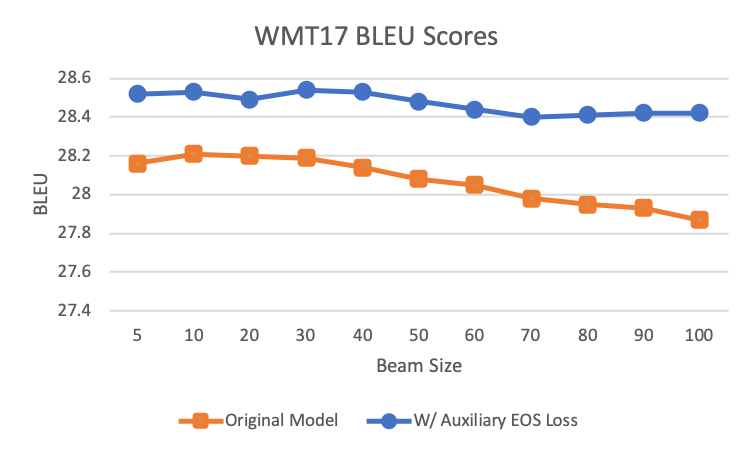}
    \caption{Results of the original model and the model with auxiliary EOS loss ($\gamma=1$) with different beam sizes. We can find the model trained with the auxiliary EOS loss is more robust to the different beam sizes compared with the original model.}
    \label{fig:aux_beam_mt17}
\end{figure}

\begin{figure}[h]
    \centering
    \includegraphics[width=0.4\textwidth]{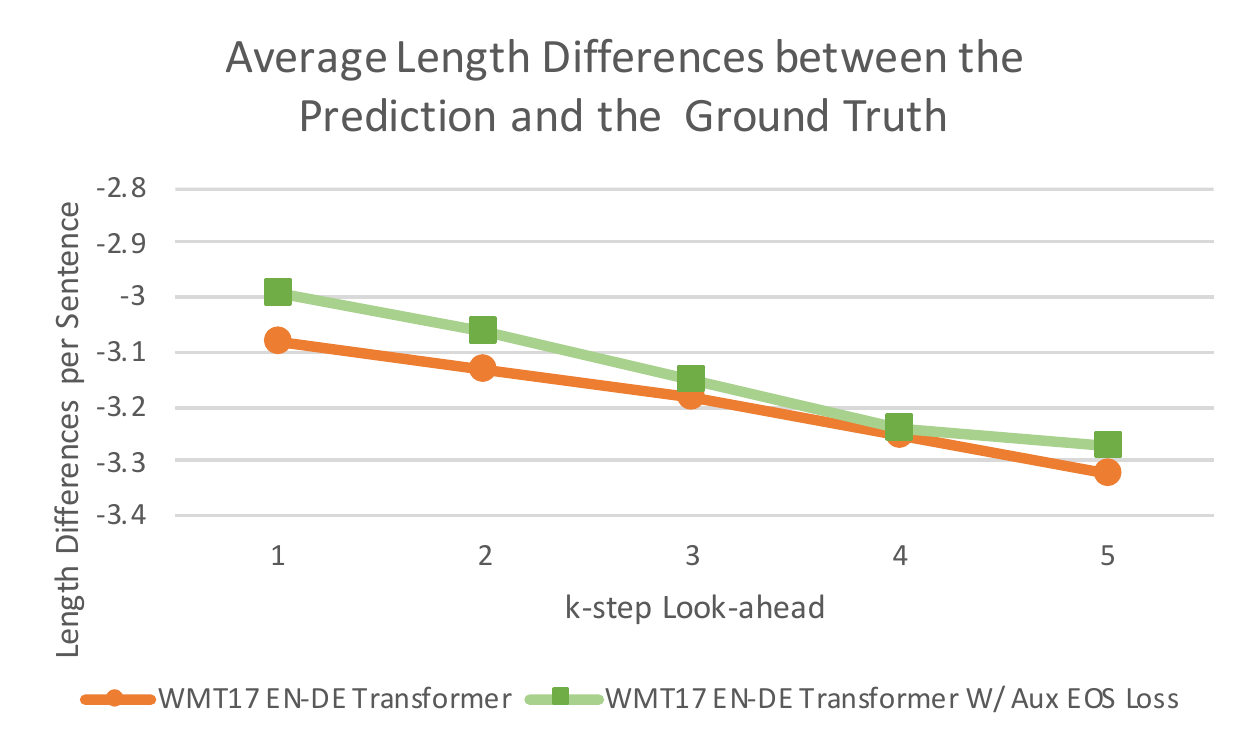}
    \caption{Average length Difference Results of the original model and the model with auxiliary EOS loss ($\gamma=1$) with different LA steps. We can find the model trained with the auxiliary EOS loss can predict longer sentences compared with the original model.}
    \label{fig:aux_len_mt17}
\end{figure}

In this experiment, we add the auxiliary EOS loss into the transformer models. We set the $\gamma$ in \ref{eq: eos} equals to 0.0 (the original model), 0.25, 0.5 0.75,1.0 and 1.25. The results are shown in Table \ref{tab:aux_la_mt17}. Surprisingly, the EOS loss consistently enhances the models with the greedy search strategy. Moreover, the model trained with the auxiliary loss is more robust to the longer look-ahead steps with the auxiliary weights smaller than 1.25. In our setting, we get the best results when we set $\gamma$ equals to one. Furthermore, we compare the auxiliary EOS loss model ($\gamma=1$) with the original model with the beam search strategy. The beam search results are shown in Figure \ref{fig:aux_beam_mt17}. Our results demonstrate that the model trained with the auxiliary EOS loss surpassed the original model with a significant margin. Moreover, unlike the original model, the auxiliary EOS model is more robust to large beam width settings. In addition, we plot the average length difference results of the original model and the model with the auxiliary loss in Figure \ref{fig:aux_len_mt17}. The average length difference results show that training with the auxiliary EOS loss ($\gamma=1$) encourage the model to predict longer sequence compared with the original model.

\section{Conclusion and Future Work}
Working on the decoding strategy can help researchers pinpoint the problems of the decoders and further improve the models by the diagnosing the errors.
Our work is an example.
We investigate the decoders using our proposed look-ahead module and then fix the overestimated EOS problem.
In the look-ahead experiments, we find the look-ahead module is able to improve on some easier datasets but less effective on a more difficult dataset, WMT14.
Our analysis suggests that the overestimated EOS probability is one of the issues and we can alleviate the problem by training the model with the auxiliary EOS loss.
There are still other feasible approaches to solve the EOS problem and integrating our proposed look-ahead model.
One of the possible ways is building an external classification network to predict the EOS at each time step instead of treating the EOS as one of the vocabulary tokens.
Another approach is incorporating the look-ahead module into the training stage and calculating the auxiliary loss using the information provided by the look-ahead module.
It is also very promising to combine the look-ahead module with beam search.
We hope this work can encourage other search strategies in the decoder and other methods to analyze the model errors in the future.

\bibliographystyle{named}
\bibliography{ijcai20}

\begin{thebibliography}{}

\bibitem[\protect\citeauthoryear{Antol \bgroup \em et al.\egroup
  }{2015}]{DBLP:conf/iccv/AntolALMBZP15}
Stanislaw Antol, Aishwarya Agrawal, Jiasen Lu, Margaret Mitchell, Dhruv Batra,
  C.~Lawrence Zitnick, and Devi Parikh.
\newblock {VQA:} visual question answering.
\newblock In {\em 2015 {IEEE} International Conference on Computer Vision,
  {ICCV} 2015, Santiago, Chile, December 7-13, 2015}, pages 2425--2433, 2015.

\bibitem[\protect\citeauthoryear{Bahdanau \bgroup \em et al.\egroup
  }{2015}]{DBLP:journals/corr/BahdanauCB14}
Dzmitry Bahdanau, Kyunghyun Cho, and Yoshua Bengio.
\newblock Neural machine translation by jointly learning to align and
  translate.
\newblock In {\em 3rd International Conference on Learning Representations,
  {ICLR} 2015, San Diego, CA, USA, May 7-9, 2015, Conference Track
  Proceedings}, 2015.

\bibitem[\protect\citeauthoryear{Chorowski and
  Jaitly}{2017}]{DBLP:conf/interspeech/ChorowskiJ17}
Jan Chorowski and Navdeep Jaitly.
\newblock Towards better decoding and language model integration in sequence to
  sequence models.
\newblock In {\em Interspeech 2017, 18th Annual Conference of the International
  Speech Communication Association, Stockholm, Sweden, August 20-24, 2017},
  pages 523--527, 2017.

\bibitem[\protect\citeauthoryear{Deng \bgroup \em et al.\egroup
  }{2017}]{latexocr}
Yuntian Deng, Anssi Kanervisto, Jeffrey Ling, and Alexander~M. Rush.
\newblock Image-to-markup generation with coarse-to-fine attention.
\newblock In {\em Proceedings of the 34th International Conference on Machine
  Learning, {ICML} 2017, Sydney, NSW, Australia, 6-11 August 2017}, pages
  980--989, 2017.

\bibitem[\protect\citeauthoryear{{Elliott} \bgroup \em et al.\egroup
  }{2016}]{multi30k}
D.~{Elliott}, S.~{Frank}, K.~{Sima'an}, and L.~{Specia}.
\newblock Multi30k: Multilingual english-german image descriptions.
\newblock pages 70--74, 2016.

\bibitem[\protect\citeauthoryear{Gehring \bgroup \em et al.\egroup
  }{2017}]{DBLP:conf/acl/GehringAGD17}
Jonas Gehring, Michael Auli, David Grangier, and Yann~N. Dauphin.
\newblock A convolutional encoder model for neural machine translation.
\newblock In {\em Proceedings of the 55th Annual Meeting of the Association for
  Computational Linguistics, {ACL} 2017, Vancouver, Canada, July 30 - August 4,
  Volume 1: Long Papers}, pages 123--135, 2017.

\bibitem[\protect\citeauthoryear{Graves}{2012}]{rnn}
Alex Graves.
\newblock {\em Supervised Sequence Labelling with Recurrent Neural Networks},
  volume 385 of {\em Studies in Computational Intelligence}.
\newblock Springer, 2012.

\bibitem[\protect\citeauthoryear{Hochreiter and Schmidhuber}{1997}]{lstm}
Sepp Hochreiter and J{\"{u}}rgen Schmidhuber.
\newblock Long short-term memory.
\newblock {\em Neural Computation}, 9(8):1735--1780, 1997.

\bibitem[\protect\citeauthoryear{{Klein} \bgroup \em et al.\egroup
  }{}]{2017opennmt}
G.~{Klein}, Y.~{Kim}, Y.~{Deng}, J.~{Senellart}, and A.~M. {Rush}.
\newblock {OpenNMT: Open-Source Toolkit for Neural Machine Translation}.
\newblock {\em ArXiv e-prints}.

\bibitem[\protect\citeauthoryear{Lei \bgroup \em et al.\egroup
  }{2018}]{DBLP:conf/emnlp/LeiYBB18}
Jie Lei, Licheng Yu, Mohit Bansal, and Tamara~L. Berg.
\newblock {TVQA:} localized, compositional video question answering.
\newblock In {\em Proceedings of the 2018 Conference on Empirical Methods in
  Natural Language Processing, Brussels, Belgium, October 31 - November 4,
  2018}, pages 1369--1379, 2018.

\bibitem[\protect\citeauthoryear{Li \bgroup \em et al.\egroup
  }{2017}]{DBLP:conf/emnlp/LiMSJRJ17}
Jiwei Li, Will Monroe, Tianlin Shi, S{\'{e}}bastien Jean, Alan Ritter, and Dan
  Jurafsky.
\newblock Adversarial learning for neural dialogue generation.
\newblock In {\em Proceedings of the 2017 Conference on Empirical Methods in
  Natural Language Processing, {EMNLP} 2017, Copenhagen, Denmark, September
  9-11, 2017}, pages 2157--2169, 2017.

\bibitem[\protect\citeauthoryear{Norouzi \bgroup \em et al.\egroup
  }{2016}]{rewardmle}
Mohammad Norouzi, Samy Bengio, zhifeng Chen, Navdeep Jaitly, Mike Schuster,
  Yonghui Wu, and Dale Schuurmans.
\newblock Reward augmented maximum likelihood for neural structured prediction.
\newblock In D.~D. Lee, M.~Sugiyama, U.~V. Luxburg, I.~Guyon, and R.~Garnett,
  editors, {\em Advances in Neural Information Processing Systems 29}, pages
  1723--1731. Curran Associates, Inc., 2016.

\bibitem[\protect\citeauthoryear{Post}{2018}]{post-2018-call}
Matt Post.
\newblock A call for clarity in reporting {BLEU} scores.
\newblock In {\em Proceedings of the Third Conference on Machine Translation:
  Research Papers}, pages 186--191, Brussels, Belgium, October 2018.
  Association for Computational Linguistics.

\bibitem[\protect\citeauthoryear{Sabour \bgroup \em et al.\egroup }{2019}]{OCD}
Sara Sabour, William Chan, and Mohammad Norouzi.
\newblock Optimal completion distillation for sequence learning.
\newblock In {\em 7th International Conference on Learning Representations,
  {ICLR} 2019, New Orleans, LA, USA, May 6-9, 2019}, 2019.

\bibitem[\protect\citeauthoryear{Silver \bgroup \em et al.\egroup
  }{2017}]{alphago}
David Silver, Julian Schrittwieser, Karen Simonyan, Ioannis Antonoglou, Aja
  Huang, Arthur Guez, Thomas Hubert, Lucas Baker, Matthew Lai, Adrian Bolton,
  Yutian Chen, Timothy Lillicrap, Fan Hui, Laurent Sifre, George Driessche,
  Thore Graepel, and Demis Hassabis.
\newblock Mastering the game of go without human knowledge.
\newblock {\em Nature}, 550:354--359, 10 2017.

\bibitem[\protect\citeauthoryear{Stahlberg and Byrne}{2019}]{nmterror}
Felix Stahlberg and Bill Byrne.
\newblock On {NMT} search errors and model errors: Cat got your tongue?
\newblock In {\em Proceedings of the 2019 Conference on Empirical Methods in
  Natural Language Processing and the 9th International Joint Conference on
  Natural Language Processing (EMNLP-IJCNLP)}, pages 3354--3360, Hong Kong,
  China, November 2019. Association for Computational Linguistics.

\bibitem[\protect\citeauthoryear{Vaswani \bgroup \em et al.\egroup
  }{2017}]{transformer}
Ashish Vaswani, Noam Shazeer, Niki Parmar, Jakob Uszkoreit, Llion Jones,
  Aidan~N. Gomez, Lukasz Kaiser, and Illia Polosukhin.
\newblock Attention is all you need.
\newblock In {\em Advances in Neural Information Processing Systems 30: Annual
  Conference on Neural Information Processing Systems 2017, 4-9 December 2017,
  Long Beach, CA, {USA}}, pages 5998--6008, 2017.

\bibitem[\protect\citeauthoryear{Vinyals \bgroup \em et al.\egroup
  }{2015}]{DBLP:conf/cvpr/VinyalsTBE15}
Oriol Vinyals, Alexander Toshev, Samy Bengio, and Dumitru Erhan.
\newblock Show and tell: {A} neural image caption generator.
\newblock In {\em {IEEE} Conference on Computer Vision and Pattern Recognition,
  {CVPR} 2015, Boston, MA, USA, June 7-12, 2015}, pages 3156--3164, 2015.

\bibitem[\protect\citeauthoryear{Wiseman and
  Rush}{2016}]{DBLP:conf/emnlp/WisemanR16}
Sam Wiseman and Alexander~M. Rush.
\newblock Sequence-to-sequence learning as beam-search optimization.
\newblock In {\em Proceedings of the 2016 Conference on Empirical Methods in
  Natural Language Processing, {EMNLP} 2016, Austin, Texas, USA, November 1-4,
  2016}, pages 1296--1306, 2016.

\bibitem[\protect\citeauthoryear{Zhou \bgroup \em et al.\egroup
  }{2018}]{DBLP:conf/cvpr/ZhouZCSX18}
Luowei Zhou, Yingbo Zhou, Jason~J. Corso, Richard Socher, and Caiming Xiong.
\newblock End-to-end dense video captioning with masked transformer.
\newblock In {\em 2018 {IEEE} Conference on Computer Vision and Pattern
  Recognition, {CVPR} 2018, Salt Lake City, UT, USA, June 18-22, 2018}, pages
  8739--8748, 2018.

\end{thebibliography}

\end{document}